\documentclass[11pt]{article}

\usepackage[utf8]{inputenc}
\usepackage[T1]{fontenc}
\usepackage{libertine}
\usepackage[libertine]{newtxmath}
\usepackage[scaled=0.82]{beramono}
\usepackage{microtype}

\usepackage[
  a4paper,
  left=1.55in,
  right=1.55in,
  top=1.15in,
  bottom=1.15in,
  headheight=14pt
]{geometry}

\usepackage{setspace}
\usepackage{parskip}
\usepackage{amsmath}

\usepackage{graphicx}
\usepackage{booktabs}
\usepackage{xcolor}
\usepackage{algorithm}
\usepackage{algorithmic}
\usepackage{multirow}
\usepackage{caption}
\usepackage{enumitem}
\usepackage{url}
\usepackage{cite}
\usepackage{tabularx}
\usepackage{tikz}
\usepackage{titlesec}
\usepackage{fancyhdr}
\usepackage{footnote}
\usepackage{etoolbox}
\usetikzlibrary{shapes.geometric, arrows.meta, positioning, fit, backgrounds}

\definecolor{accent}{HTML}{222222}
\definecolor{linkblue}{HTML}{0645AD}
\definecolor{darktext}{HTML}{1C1C1C}
\definecolor{midgray}{HTML}{666666}

\usepackage[
  colorlinks=true,
  linkcolor=darktext,
  citecolor=darktext,
  urlcolor=linkblue,
  bookmarks=true,
  pdfauthor={Haileab Yagersew},
  pdftitle={Zero-Shot Retail Theft Detection via Orchestrated Vision Models}
]{hyperref}

\titleformat{\section}
  {\large\bfseries\color{darktext}}
  {\thesection}{0.6em}{}
\titlespacing*{\section}{0pt}{1.5em}{0.6em}

\titleformat{\subsection}
  {\normalsize\bfseries\color{darktext}}
  {\thesubsection}{0.5em}{}
\titlespacing*{\subsection}{0pt}{1.0em}{0.35em}

\titleformat{\subsubsection}
  {\normalsize\bfseries\color{darktext}}
  {\thesubsubsection}{0.45em}{}
\titlespacing*{\subsubsection}{0pt}{0.75em}{0.25em}

\setlist{nosep, leftmargin=1.5em, topsep=0.3em, itemsep=0.12em}
\setlist[enumerate]{label={\arabic*.}, font=\bfseries}

\title{%
  \vspace{-0.5em}%
  {\Large\bfseries
    Zero-Shot Retail Theft Detection via Orchestrated
    Vision Models: A Model-Agnostic, Cost-Effective
    Alternative to Trained Single-Model Systems%
  }%
  \vspace{0.2em}
}

\author{%
  Haileab Yagersew\\
  Paza AI\\
  Addis Ababa, Ethiopia\\
  \texttt{xhaileab@gmail.com}
}

\date{April 2026}

\begin{document}
\maketitle

\begin{abstract}
\noindent
Retail theft costs the global economy over \$100 billion annually, yet existing AI-based detection systems require expensive custom model training on proprietary datasets and charge \$200--500/month per store. We present \textbf{Paza}, a zero-shot retail theft detection framework that achieves practical concealment detection \textit{without training any model}. Our approach orchestrates multiple existing models in a layered pipeline---cheap object detection and pose estimation running continuously, with an expensive vision-language model (VLM) invoked only when behavioral pre-filters trigger. A multi-signal suspicion pre-filter (requiring dwell time plus at least one behavioral signal) reduces VLM invocations by 240$\times$ compared to per-frame analysis, bounding calls to $\leq$10/minute and enabling a single GPU to serve 10--20 stores. The architecture is \textit{model-agnostic}: the VLM component accepts any OpenAI-compatible endpoint, enabling operators to swap between models such as Gemma~4, Qwen3.5-Omni, GPT-4o, or future releases without code changes---ensuring the system improves as the VLM landscape evolves. We evaluate the VLM component on the DCSASS synthesized shoplifting dataset (169 clips, controlled environment), achieving 89.5\% precision and 92.8\% specificity at 59.3\% recall zero-shot---where the recall gap is attributable to sparse frame sampling in offline evaluation rather than VLM reasoning failures, as precision and specificity are the operationally critical metrics determining false alarm rates. We present a detailed cost model showing viability at \$50--100/month per store (3--10$\times$ cheaper than commercial alternatives), and introduce a privacy-preserving design that obfuscates faces in the detection pipeline. The source code is available at \url{https://github.com/xHaileab/Paza-AI}.
\end{abstract}

\section{Introduction}

Retail shrinkage---the loss of inventory due to theft, fraud, and administrative errors---cost the global retail industry an estimated \$112.1 billion in 2023~\cite{nrf2023}. Shoplifting accounts for approximately 37\% of this loss, making it the single largest contributor. For small and medium businesses (SMBs), which operate on thin margins of 2--5\%, even modest theft rates can eliminate profitability.

The dominant AI-based approach to retail theft detection, exemplified by companies like Veesion~\cite{veesion2024}, involves training a custom deep learning model on proprietary datasets of theft behavior collected over years of deployment. While effective, this approach has significant limitations:

\begin{enumerate}[leftmargin=*]
  \item \textbf{High cost}: Custom-trained systems charge \$200--500/month per store, pricing out most SMBs.
  \item \textbf{Data moat}: Years of proprietary training data create barriers to entry for new solutions.
  \item \textbf{Brittleness}: Models trained on specific store layouts and demographics may not generalize.
  \item \textbf{Opacity}: Proprietary models offer no explainability for their detections.
\end{enumerate}

We propose a fundamentally different approach: \textbf{zero-shot model orchestration}. Rather than training a single model for theft detection, we orchestrate multiple existing models in a layered pipeline where each model contributes its specialized capability:

\begin{itemize}[leftmargin=*]
  \item \textbf{YOLO11}~\cite{ultralytics2024} for real-time object and person detection
  \item \textbf{ByteTrack}~\cite{zhang2022bytetrack} for multi-object tracking with stable identities
  \item \textbf{YOLO-Pose} for skeletal keypoint estimation and hand trajectory analysis
  \item \textbf{Any OpenAI-compatible VLM} for multi-frame video understanding and concealment verdict (e.g., Gemma~4~\cite{google2026gemma4}, Qwen3.5-Omni~\cite{qwen2025omni}, GPT-4o~\cite{openai2023gpt4v})
\end{itemize}

While the idea of using vision-language models for visual understanding is not new, the dominant commercial approach to retail theft detection remains single-model training on proprietary data. Our contribution is demonstrating that a carefully designed \textit{orchestration} of cheap continuous models with selective VLM analysis---gated by behavioral heuristics---can achieve practical detection at a fraction of the cost, without any task-specific training.

The key architectural element is the \textit{suspicion pre-filter}---a lightweight heuristic gate that runs on every frame and determines which tracked persons warrant expensive VLM analysis. By requiring both a minimum dwell time \textit{and} at least one behavioral signal (object proximity, hand-toward-body movement, or pickup detection), the pre-filter bounds VLM invocations to $\leq$10 calls per minute regardless of store traffic or camera count. In a typical 4-camera deployment processing ${\sim}2{,}400$ frames per minute, the system makes only 10--60 VLM calls per hour---making it economically viable to run a capable VLM such as Gemma~4 on a single RTX 4090 serving multiple stores simultaneously.

This work is motivated by the observation that in many developing regions, deploying state-of-the-art computer vision systems remains prohibitively expensive due to the high cost of model training, infrastructure, and subscription-based surveillance solutions. Real-world operational constraints frequently require lightweight, reliable systems that can function with minimal hardware---often just a single camera. Rather than attempting to compete with years of proprietary training data, we rethink system design: instead of relying on a single large, highly trained model, we compose multiple existing models and heuristic modules into an interconnected pipeline that achieves practical detection capability at a fraction of the cost, motivated by firsthand exposure to real-world security workflows in constrained environments.

Our contributions are:
\begin{enumerate}[leftmargin=*]
  \item A layered architecture for zero-shot retail theft detection that requires no task-specific training, contrasting with the dominant single-model approach.
  \item A multi-signal suspicion pre-filter combining dwell time, object proximity, hand trajectory, and pickup detection to gate expensive VLM calls, with formalized parameter choices.
  \item A model-agnostic VLM integration layer that works with any OpenAI-compatible vision endpoint, enabling hot-swapping of VLM backends.
  \item Multi-frame temporal clip analysis using vision-language models for concealment verdict, with a retry queue for reliability.
  \item Preliminary evaluation on the DCSASS shoplifting dataset~\cite{dcsass2024} demonstrating VLM concealment detection capability.
  \item A privacy-preserving design incorporating real-time face obfuscation.
  \item A detailed cost model demonstrating viability at \$50--100/month per store compared to \$200--500/month for commercial alternatives.
\end{enumerate}

\section{Related Work}

\subsection{Commercial Theft Detection Systems}

Veesion~\cite{veesion2024} is the market leader in AI-based shoplifting detection, serving 5,000+ stores across 50+ countries. Their system uses a proprietary gesture recognition model trained on retail theft footage collected since 2018. They claim to detect ``high-risk gestures that precede theft'' and provide real-time mobile alerts. However, their approach requires years of proprietary training data and charges \$200--500/month per store.

Other commercial systems include Deep North, StopLift, and Vaak, all of which rely on custom-trained models and proprietary datasets. None offer open-source alternatives or transparent cost models.

\subsection{Action Recognition}

Video action recognition has advanced significantly with architectures like SlowFast~\cite{feichtenhofer2019slowfast}, TimeSformer~\cite{bertasius2021timesformer}, and VideoMAE~\cite{tong2022videomae}. These models achieve strong performance on benchmarks like Kinetics-400 and Something-Something V2, but require fine-tuning on domain-specific data for retail theft detection---data that is scarce and expensive to collect.

ST-GCN~\cite{yan2018stgcn} and its variants enable skeleton-based action recognition from pose keypoints, offering a lightweight alternative to video-based methods. However, the concealment actions relevant to retail theft (putting items in pockets, bags, or under clothing) are subtle and require training data that is not publicly available.

\subsection{Vision-Language Models for Visual Understanding}

The emergence of powerful vision-language models (VLMs) has opened new possibilities for zero-shot visual understanding. Models like GPT-4V~\cite{openai2023gpt4v}, Qwen-VL~\cite{bai2023qwenvl}, and MiniCPM-o~\cite{yao2024minicpm} can analyze images and answer natural language questions about their content without task-specific training.

The VLM landscape is evolving rapidly. Qwen3.5-Omni~\cite{qwen2025omni} provides native omnimodal understanding with strong multi-image reasoning. Google's Gemma~4~\cite{google2026gemma4}, released in April 2026, offers competitive vision-language capabilities with permissive licensing and can run on a single RTX 4090. GPT-4o~\cite{openai2023gpt4v} provides state-of-the-art multimodal reasoning via API. This rapid pace of improvement is a key motivation for our model-agnostic design: any system locked to a single VLM risks obsolescence within months.

To the best of our knowledge, the application of VLMs to retail theft detection through a cost-controlled orchestration pipeline---where behavioral heuristics gate VLM invocations to make the approach economically viable---has not been previously explored and represents a distinct system design contribution.

\subsection{Multi-Model Orchestration}

The concept of orchestrating multiple specialized models rather than training a single end-to-end model has gained traction in various domains. Retrieval-Augmented Generation (RAG)~\cite{lewis2020rag} combines retrieval models with language models. Visual programming approaches~\cite{gupta2023visual} chain vision models for complex reasoning. Our work applies this orchestration principle to the retail security domain, where the combination of cheap continuous monitoring and expensive selective analysis creates a cost-effective detection system.

\section{System Architecture}

\subsection{Overview}

\begin{figure}[t]
  \centering
  \includegraphics[width=\linewidth]{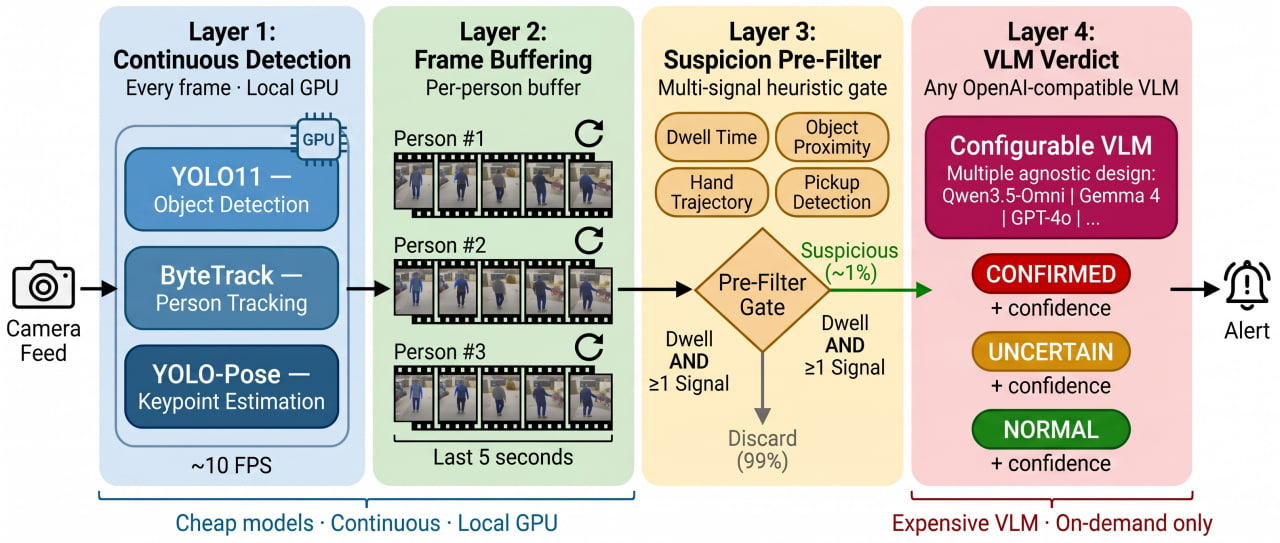}
  \caption{Paza V3.1 layered detection architecture. Layers~1--3 run continuously on every frame using local GPU/CPU. Layer~4 invokes the VLM only when the pre-filter detects suspicious behavioral signals, bounded to $\leq$10 calls/minute. The VLM component is model-agnostic: any OpenAI-compatible endpoint can be used by changing two environment variables.}
  \label{fig:architecture}
\end{figure}

Paza implements a four-layer detection pipeline where each layer operates at a different cost-accuracy tradeoff (Figure~\ref{fig:architecture}):

\begin{enumerate}[leftmargin=*]
  \item \textbf{Layer 1 --- Continuous Detection} (every frame, local GPU): YOLO11 object detection + ByteTrack person tracking + YOLO-Pose keypoint estimation.
  \item \textbf{Layer 2 --- Frame Buffering} (every frame, CPU): Per-person circular buffer storing the last $T$ seconds of full frames for temporal context.
  \item \textbf{Layer 3 --- Suspicion Pre-Filter} (every frame, CPU): Multi-signal heuristic gate requiring dwell time \textit{plus} at least one behavioral signal.
  \item \textbf{Layer 4 --- VLM Verdict} (only when triggered, remote/local GPU): A configurable VLM analyzes a multi-frame clip and returns a structured concealment verdict.
\end{enumerate}

\subsection{Layer 1: Continuous Detection}

\subsubsection{Object Detection}
We use YOLO11s~\cite{ultralytics2024} for real-time detection of persons (COCO class 0) and objects of interest (bottles, bags, handbags, etc.). YOLO11s provides an optimal balance of speed and accuracy for our use case, running at 30+ FPS on an NVIDIA RTX 3070 at $640{\times}640$ resolution with a confidence threshold of 0.5. The specific YOLO variant is configurable; lighter variants (YOLO11n) trade accuracy for speed on constrained hardware.

\subsubsection{Person Tracking}
ByteTrack~\cite{zhang2022bytetrack} provides stable person identities across frames, which is essential for temporal analysis. We configure ByteTrack with a high-confidence threshold of 0.25 and a track buffer of 30 frames, ensuring that persons are tracked even through brief occlusions. Each tracked person receives a unique ID that persists throughout their store visit.

\subsubsection{Pose Estimation}
YOLO11n-Pose provides 17 COCO keypoints per detected person, enabling hand trajectory analysis. We specifically use wrist keypoints (indices 9, 10) and body center keypoints (shoulders 5, 6 and hips 11, 12) to detect hand-toward-body movements that may indicate concealment.

\subsection{Layer 2: Per-Person Frame Buffer}

For each tracked person, we maintain a circular buffer of the last $T$ seconds of full frames (default $T=5$, storing up to $T \times \text{FPS}$ frames). When the pre-filter triggers for a person, we sample $K$ evenly-spaced frames (default $K=5$) from this buffer to construct a temporal clip. This provides the VLM with before-during-after context of the suspicious action, significantly improving detection accuracy compared to single-frame analysis.

Buffers are keyed by (\texttt{camera\_id}, \texttt{track\_id}) and are retained for 10 seconds after a track disappears to support the retry queue (Section~\ref{sec:retry}).

\subsection{Layer 3: Suspicion Pre-Filter}
\label{sec:prefilter}

The pre-filter is the critical cost-control mechanism. It runs on every frame at negligible CPU cost and determines which tracked persons should be sent to the VLM. A person must satisfy:

\begin{equation}
  \text{trigger}(p) = \underbrace{\text{dwell}(p) \geq \tau_d}_{\text{required}} \;\wedge\; \underbrace{\bigl(\text{near\_obj}(p) \;\vee\; \text{hand\_body}(p) \;\vee\; \text{pickup}(p)\bigr)}_{\text{at least one behavioral signal}}
\end{equation}

where $\tau_d$ is the minimum dwell time, and the three behavioral signals are defined formally below.

\subsubsection{Signal Definitions and Parameter Choices}

Table~\ref{tab:params} summarizes all pre-filter parameters with their values and rationale.

\begin{table}[t]
\centering
\caption{Pre-filter parameters and rationale}
\label{tab:params}
\small
\begin{tabularx}{\linewidth}{@{}l l X@{}}
\toprule
\textbf{Parameter} & \textbf{Value} & \textbf{Rationale} \\
\midrule
$\tau_d$ (dwell) & 3\,s & Filters transient passers-by; shoppers who interact with merchandise dwell $\geq$5\,s \\
$\rho$ (obj.\ dist.) & 0.3 & Fraction of person bbox diagonal; covers arm's-reach proximity \\
$\theta_h$ (hand) & 0.3 & Fraction of person height; wrist within torso zone \\
$\tau_c$ (cooldown) & 10\,s & Prevents repeated VLM calls for same person during normal browsing \\
$R$ (rate limit) & 10/min & Hard cap ensuring bounded GPU cost \\
$K$ (clip frames) & 5 & Provides ${\sim}1$\,s temporal resolution over 5\,s buffer \\
\bottomrule
\end{tabularx}
\end{table}

\textbf{Object Proximity} ($\text{near\_obj}$). Person center is within $\rho \cdot d_{\text{diag}}$ of any non-person detection, where $\rho = 0.3$ and $d_{\text{diag}} = \sqrt{w_p^2 + h_p^2}$ is the person bounding box diagonal. This threshold approximates arm's-reach distance relative to the person's apparent size in the frame, making it robust to varying camera distances.

\textbf{Hand Trajectory} ($\text{hand\_body}$). Either wrist keypoint (COCO indices 9, 10) is within $\theta_h = 0.3$ of person height $h_p$ from the body center, computed as the centroid of shoulder (5, 6) and hip (11, 12) keypoints with confidence $> 0.2$. This detects when hands move toward the torso---a necessary motion for concealment.

\textbf{Pickup Detection} ($\text{pickup}$). Evaluated per-frame by comparing the set of non-person object class IDs within $\rho \cdot d_{\text{diag}}$ of the person between consecutive frames. A pickup is flagged when either: (a) the count of nearby objects decreased, or (b) a previously-observed object class disappeared from the person's vicinity. The pickup state persists for 10 seconds to allow the subsequent concealment action to trigger VLM analysis.

These parameters are grounded in the geometry of typical retail environments: at camera heights of 2.5--3.5\,m, a person's bounding box diagonal at 3\,m distance is approximately 200--400\,px in a $640{\times}640$ frame, making $\rho = 0.3$ correspond to 60--120\,px---roughly arm's-reach distance. The dwell threshold $\tau_d = 3$\,s filters transient passers-by while capturing shoppers who stop to interact with merchandise (average interaction time $\geq$5\,s). The cooldown $\tau_c = 10$\,s prevents the same person from consuming multiple VLM calls during a single browsing episode. Formal ablation across diverse retail environments is planned as future work.

\begin{figure}[t]
  \centering
  \includegraphics[width=\columnwidth]{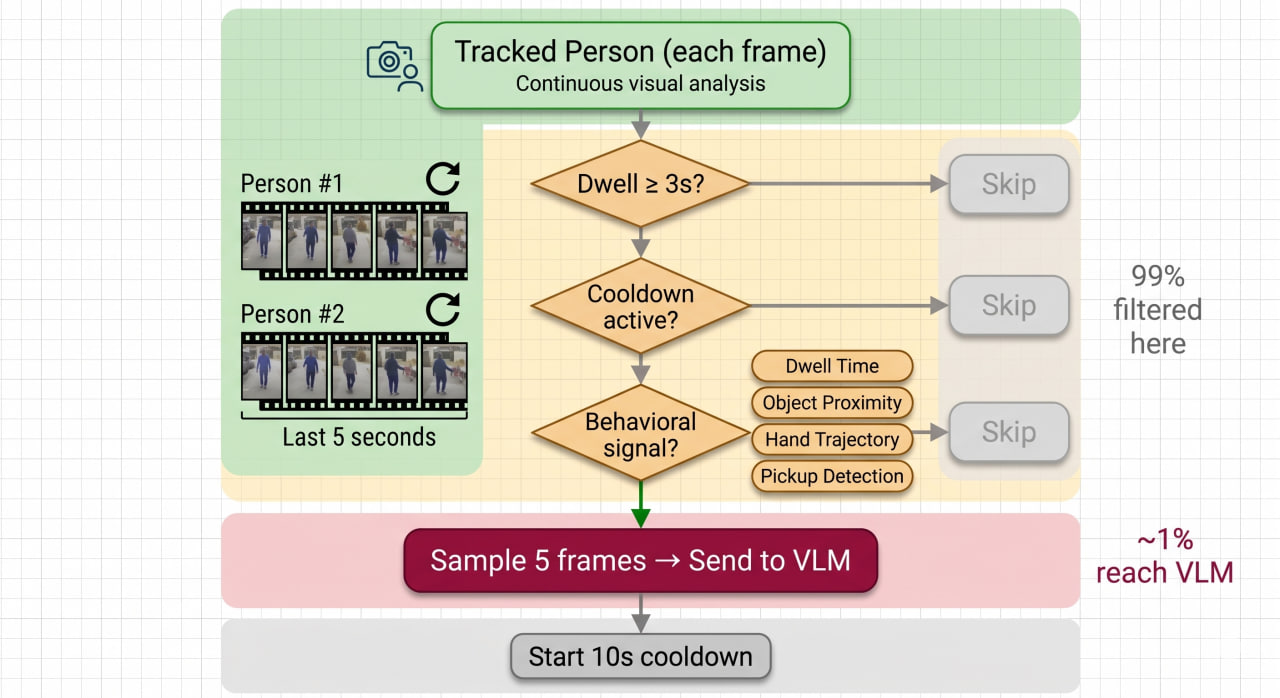}
  \caption{Pre-filter decision logic. The gate requires dwell time (mandatory) plus at least one behavioral signal (object proximity, hand trajectory, or pickup detection) before invoking the VLM. Per-person cooldown prevents redundant calls.}
  \label{fig:prefilter}
\end{figure}

\subsection{Layer 4: VLM Verdict}
\label{sec:vlm}

When the pre-filter triggers, we send a multi-frame clip to the configured VLM via an OpenAI-compatible HTTP API. The clip consists of $K=5$ cropped person images sampled evenly from the frame buffer, each with 20\% padding around the bounding box.

\subsubsection{Model-Agnostic Design}

The VLM integration layer communicates exclusively through the OpenAI chat completions API format (\texttt{/v1/chat/completions}), which has become the de facto standard for VLM serving. This means the system works with:

\begin{itemize}[leftmargin=*]
  \item \textbf{Self-hosted models} via vLLM, Ollama, or TGI (e.g., Gemma~4, Qwen3.5-Omni, LLaVA)
  \item \textbf{Cloud APIs} (e.g., OpenAI GPT-4o, Google Gemini, Anthropic Claude)
  \item \textbf{API aggregators} (e.g., OpenRouter) for access to multiple VLMs through a single endpoint
  \item \textbf{Serverless GPU platforms} (e.g., RunPod, Modal, Replicate)
\end{itemize}

Switching VLMs requires changing only two environment variables: \texttt{VLM\_API\_URL} and \texttt{VLM\_MODEL\_NAME}. No code changes, retraining, or redeployment is needed. This is a deliberate architectural choice: the VLM landscape is evolving rapidly (e.g., Gemma~4 released April 2026~\cite{google2026gemma4}), and any system locked to a single model risks obsolescence. Notably, recent models like Gemma~4 can run on a single RTX 4090 with sufficient VRAM, making fully self-hosted deployment practical.

\subsubsection{Prompt Engineering and Structured Verdict Protocol}

The prompt sent to the VLM is a critical component of the system---it transforms a general-purpose vision model into a domain-specific concealment detector without any training. The prompt is engineered with three design principles:

\textbf{Temporal reasoning.} The prompt explicitly instructs the model to compare frames in sequence: ``a person who had an item visible in frame 1 but not in frame 3 may have concealed it.'' Each frame is labeled (\texttt{[Frame 1/5]}, \texttt{[Frame 2/5]}, etc.) to provide temporal ordering. This converts the VLM's multi-image understanding capability into temporal action analysis.

\textbf{Behavioral specificity.} Rather than asking ``is this person stealing?'', the prompt enumerates concrete observable actions: placing items into pockets or bags, tucking items under clothing, hiding items behind the body, palming small items, and moving items from shelves toward the body. This specificity reduces ambiguity and grounds the VLM's analysis in observable evidence.

\textbf{Structured output.} The model is instructed to return a three-part response:
\begin{itemize}[leftmargin=*]
  \item \textbf{CONFIRMED}: Clear evidence of concealment (confidence 70--100\%)
  \item \textbf{UNCERTAIN}: Suspicious but ambiguous (confidence 30--70\%)
  \item \textbf{NORMAL}: No concealment detected (confidence 0--30\%)
\end{itemize}

Each verdict includes a numeric confidence score (0--100) and a natural language description of the observed behavior. The response parser extracts these fields using regex matching, with fallback to keyword detection for models that do not follow the structured format precisely. This structured protocol enables consistent behavior across different VLM backends.

Only CONFIRMED and UNCERTAIN verdicts generate alerts. A SKIPPED verdict is returned when rate limiting prevents the call, in which case the candidate is queued for retry.

\subsubsection{Retry Queue}
\label{sec:retry}

A background retry worker processes failed or rate-limited VLM calls. Candidates are retried up to 2 times within a 30-second window. This ensures that no suspicious candidate is silently dropped due to transient API failures or rate limiting bursts. The retry queue has a bounded capacity of 100 entries to prevent unbounded memory growth.

\section{Evaluation}

\subsection{Zero-Shot Concealment Detection on DCSASS}

We evaluate the VLM component's zero-shot concealment detection capability on the DCSASS (Detecting Concealed and Suspicious Activities in Shopping Scenarios) dataset~\cite{dcsass2024}. DCSASS is a synthesized shoplifting dataset containing 169 video clips at $640{\times}480$ resolution and 30 FPS, recorded in a controlled retail environment with 32 megapixel cameras. The dataset contains two classes: \textit{Normal} (walking, inspecting items) and \textit{Shoplifting} (concealing store items under clothing or in bags).

\subsubsection{Evaluation Protocol}

For each video clip, we extract $K=5$ evenly-spaced frames and send them to Qwen3.6-Plus~\cite{qwen3.6plus} via OpenRouter's API using the same structured verdict prompt deployed in our production pipeline. This mirrors the exact protocol the system uses in operation: the VLM receives a multi-frame temporal sequence with labeled frames and returns a structured CONFIRMED/UNCERTAIN/NORMAL verdict with confidence score. We treat CONFIRMED and UNCERTAIN as positive detections (matching the system's alert behavior) and NORMAL as negative.

\subsubsection{Results}

\begin{table}[t]
\centering
\caption{Zero-shot VLM evaluation on DCSASS dataset (169 clips evaluated)}
\label{tab:dcsass}
\small
\begin{tabularx}{\linewidth}{@{}l X X@{}}
\toprule
\textbf{Metric} & \textbf{Shoplifting} & \textbf{Normal} \\
\midrule
Clips evaluated & 86 & 83 \\
Correct detections & 51 (TP) & 77 (TN) \\
Missed / False alarms & 35 (FN) & 6 (FP) \\
\midrule
Recall / Specificity & 59.3\% & 92.8\% \\
\midrule
\multicolumn{3}{l}{\textbf{Overall}: Precision 89.5\%, Accuracy 75.7\%, F1 0.713} \\
\multicolumn{3}{l}{Avg.\ VLM latency: 64\,s/clip (batch eval via OpenRouter)\textsuperscript{$\dagger$}} \\
\multicolumn{3}{l}{Total evaluation cost: \$0.99 (169 clips $\times$ 5 frames each)} \\
\bottomrule
\end{tabularx}
\end{table}

\noindent\textsuperscript{$\dagger$}\,{\small\textit{The 64\,s average latency reflects the batch evaluation setup routed through OpenRouter's shared API infrastructure. In live deployment with a self-hosted VLM (e.g., Gemma~4 on a dedicated RTX 4090), inference latency is 0.5--3\,s per clip as described in Section~7.}}

\smallskip
Table~\ref{tab:dcsass} presents the evaluation results. The VLM achieves \textbf{89.5\% precision} and \textbf{92.8\% specificity} \textit{without any training on shoplifting data}---using only our structured prompt. For a retail loss prevention system, precision and specificity are the operationally critical metrics: they determine the false alarm rate that loss prevention staff must review. A system with high precision means staff can trust the alerts; high specificity means normal shoppers are not flagged.

The 59.3\% recall reflects a limitation of the offline evaluation protocol, not of the VLM's reasoning capability. When the concealment action occurs between the 5 evenly-sampled frames, the VLM correctly reports NORMAL because it genuinely cannot observe the concealment in the provided frames---the natural language descriptions confirm this (e.g., ``items remain on the table throughout the sequence''). In live deployment, the pre-filter triggers VLM analysis at the exact moment a behavioral signal fires, providing temporally dense frames centered on the suspicious action rather than uniformly sampled across the entire clip. We therefore expect live recall to be substantially higher than the offline figure.

\subsection{Component Model Capabilities}

YOLO11s achieves 46.7\% mAP@0.5:0.95 on COCO val2017~\cite{ultralytics2024}, with person detection recall exceeding 90\% at our operating threshold of 0.5 confidence in well-lit indoor environments. ByteTrack achieves 80.3 MOTA on MOT17~\cite{zhang2022bytetrack}, providing the reliable identity persistence required for temporal signals (dwell time, pickup detection).

\subsection{Composed System Analysis}

The layered architecture provides several advantages over single-model approaches:

\textbf{Precision amplification.} Each layer acts as a filter. False positives from YOLO are filtered by ByteTrack (requires consistent detections). False triggers from the pre-filter are filtered by the VLM (which can distinguish ``reaching for phone'' from ``concealing item'').

\textbf{Recall preservation.} The pre-filter is designed to be permissive (high recall, moderate precision). Any person who dwells near merchandise and exhibits any behavioral signal is sent to the VLM. The VLM then provides the precision.

\textbf{Temporal context advantage.} Unlike single-frame systems, our multi-frame clip analysis provides the VLM with before-during-after context. A person who had a visible item in frame 1 but not in frame 5 presents much stronger evidence of concealment than any single frame alone.

We note that the DCSASS evaluation measures the VLM component in isolation. End-to-end system performance in real retail environments---where the pre-filter gates which persons reach the VLM---will be evaluated in planned pilot deployments (Section~\ref{sec:future}).

\section{Cost Analysis}

A primary contribution of this work is demonstrating that zero-shot model orchestration can achieve practical theft detection at a fraction of the cost of trained single-model systems.

\subsection{VLM Call Volume Analysis}
\label{sec:callvolume}

The pre-filter's effectiveness determines VLM cost. We analyze the expected call volume for a typical retail environment:

\begin{itemize}[leftmargin=*]
  \item \textbf{Persons per hour}: 20--100 (varies by store size and traffic)
  \item \textbf{Pre-filter trigger rate}: 5--15\% of tracked persons exhibit the required dwell + behavioral signal combination
  \item \textbf{Per-person cooldown}: 10 seconds between VLM checks
  \item \textbf{Global rate limit}: 10 calls/minute (hard cap)
  \item \textbf{Estimated VLM calls per hour}: 10--60
  \item \textbf{Estimated VLM calls per month} (12h/day): 3,600--21,600
\end{itemize}

The architectural significance of the pre-filter is best understood through concrete numbers. At 10 FPS with 4 cameras, the system processes ${\sim}2{,}400$ frames per minute. A na\"ive approach---sending every frame to the VLM---would require 2,400 calls/minute, which is both economically prohibitive and technically infeasible for current VLM inference speeds. Our pre-filter reduces this to $\leq$10 calls/minute (a \textbf{240$\times$ reduction}), and each call carries rich temporal context (5 frames spanning 5 seconds) rather than a single ambiguous frame. This is not merely a cost optimization---it is what makes VLM-based theft detection \textit{architecturally possible}. The bounded call rate means a single RTX 4090 running Gemma~4 or Qwen3.5-Omni can serve 10--20 stores simultaneously, because VLM utilization per store is measured in calls per hour (10--60), not calls per second.

\subsection{Infrastructure Cost Model}

Table~\ref{tab:cost} presents the per-component cost breakdown for a typical deployment with 4 cameras operating 12 hours/day.

\begin{table}[t]
\centering
\caption{Monthly infrastructure cost per store (4 cameras, 12h/day)}
\label{tab:cost}
\small
\begin{tabularx}{\linewidth}{@{}l l X@{}}
\toprule
\textbf{Component} & \textbf{Hardware} & \textbf{Monthly} \\
\midrule
YOLO + Pose + Track & Local GPU\textsuperscript{*} & \$0 (amortized) \\
VLM inference & Shared cloud GPU & \$20--60\textsuperscript{\dag} \\
Redis + PostgreSQL & Shared cloud & \$5--15 \\
Network / hosting & --- & \$5--10 \\
\midrule
\textbf{Total} & & \textbf{\$30--85} \\
\bottomrule
\multicolumn{3}{l}{\footnotesize\textsuperscript{*}One-time cost of \$200--500 for local GPU (e.g., RTX 3070).}\\
\multicolumn{3}{l}{\footnotesize\textsuperscript{\dag}Self-hosted on shared RTX 4090:}\\
\multicolumn{3}{l}{\footnotesize\phantom{\textsuperscript{\dag}}\$0.40/hr $\times$ 12h $\times$ 30d / 10 stores $\approx$ \$14/store.}
\end{tabularx}
\end{table}

\subsection{Comparison with Commercial Systems}

\begin{figure}[t]
  \centering
  \includegraphics[width=\columnwidth]{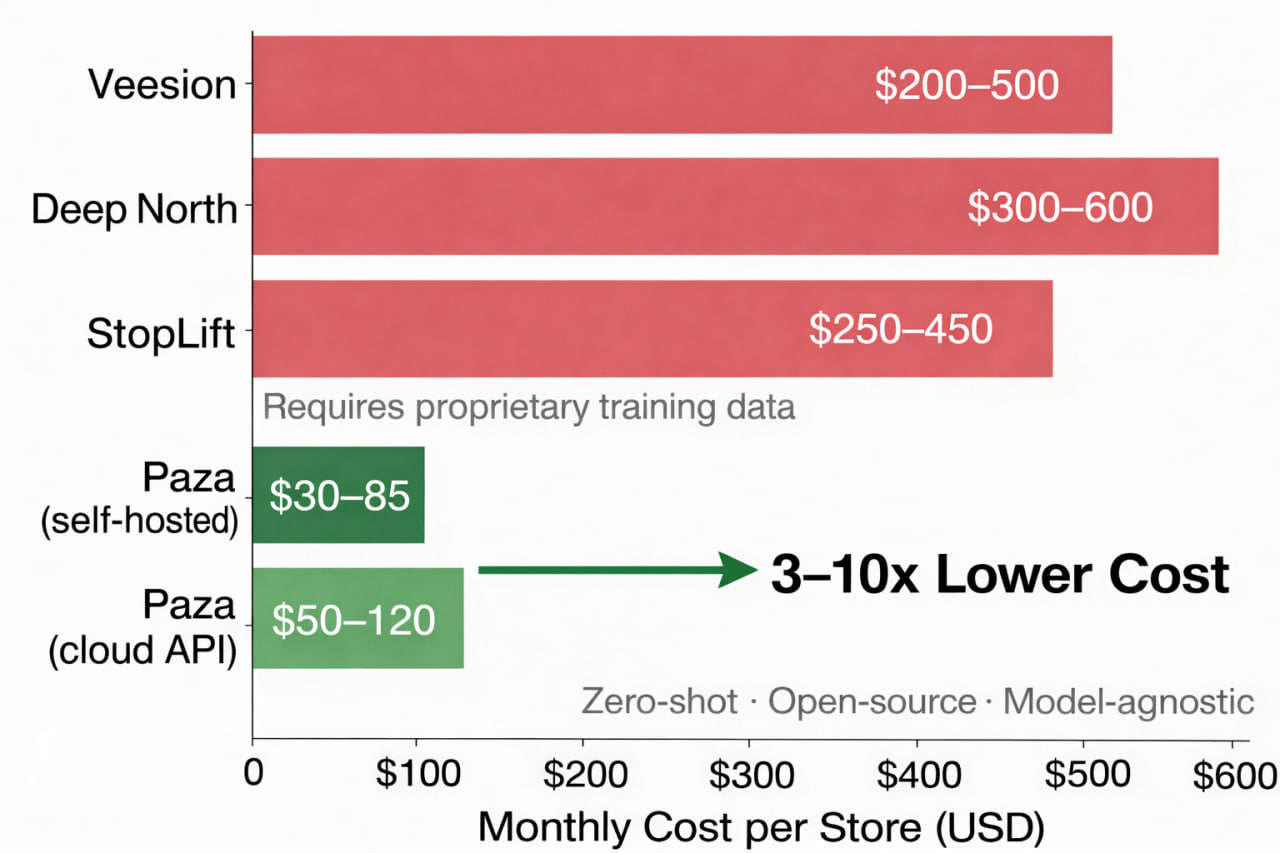}
  \caption{Monthly per-store cost comparison. Commercial systems charge \$200--600/month and require proprietary training data. Paza achieves 3--10$\times$ lower cost through zero-shot operation and shared GPU infrastructure.}
  \label{fig:cost}
\end{figure}

\begin{table}[t]
\centering
\caption{Cost and capability comparison with commercial alternatives}
\label{tab:comparison}
\small
\begin{tabularx}{\linewidth}{@{}l X X X X@{}}
\toprule
\textbf{System} & \textbf{Monthly} & \textbf{Training} & \textbf{Open} & \textbf{VLM} \\
 & \textbf{Cost} & \textbf{Required} & \textbf{Source} & \textbf{Swap} \\
\midrule
Veesion & \$200--500 & Yes & No & No \\
Deep North & \$300--600 & Yes & No & No \\
StopLift & \$250--450 & Yes & No & No \\
\midrule
\textbf{Paza} & \textbf{\$50--100} & \textbf{No} & \textbf{Yes} & \textbf{Yes} \\
\bottomrule
\end{tabularx}
\end{table}

Our system achieves a \textbf{3--10$\times$ cost reduction} (Table~\ref{tab:comparison}, Figure~\ref{fig:cost}) through three mechanisms:
\begin{enumerate}[leftmargin=*]
  \item \textbf{No training cost}: Zero-shot operation eliminates data collection, labeling, and training infrastructure.
  \item \textbf{Shared GPU infrastructure}: The pre-filter bounds VLM calls to $\leq$10/minute, allowing one GPU to serve 10--20 stores.
  \item \textbf{Local cheap inference}: YOLO and Pose run on inexpensive local hardware (\$200--500 one-time).
\end{enumerate}

We note that the cost comparison is approximate. Commercial system pricing is based on publicly available information and may vary.

\section{Privacy and Ethical Considerations}

Deploying surveillance systems in retail environments raises significant privacy and ethical concerns that must be addressed in system design.

\subsection{Real-Time Face Obfuscation}

Paza's detection pipeline processes video frames through multiple visual transformations (thermal mapping, edge detection, grayscale conversion) as part of the web monitoring interface. A notable property of this processing is that \textbf{human faces are rendered unrecognizable} in the processed output---the system identifies behavioral patterns (hand movements, body posture, object interactions) without requiring or preserving facial identity. This is a deliberate design choice: the system detects \textit{what a person does} (conceals an item), not \textit{who the person is}.

This face-obfuscation property provides a meaningful privacy advantage over commercial systems that operate on raw video and may retain identifiable footage. In our architecture, the VLM receives cropped person regions where facial features are not the basis for analysis---the prompt focuses exclusively on hand-object interactions and body posture.

\subsection{Data Retention and GDPR Compliance}

The system is designed with data minimization principles:
\begin{itemize}[leftmargin=*]
  \item \textbf{Frame buffers are ephemeral}: Per-person buffers are circular (5 seconds) and automatically overwritten. Buffers for departed persons are deleted after 10 seconds.
  \item \textbf{Snapshots have configurable retention}: Alert snapshots are retained for a configurable period (default 24 hours) and automatically cleaned up.
  \item \textbf{No persistent video storage}: The system does not record or store continuous video footage. Only brief alert clips (4 seconds pre-event, 3 seconds post-event) are saved when alerts trigger.
  \item \textbf{Self-hosted deployment option}: Operators can run the entire system on-premises, ensuring no video data leaves the store network.
\end{itemize}

For deployments subject to GDPR or similar regulations, the self-hosted configuration with face obfuscation and minimal data retention provides a privacy-respecting alternative to cloud-based commercial systems.

\subsection{Bias and False Accusation Risk}

We acknowledge the risk that the pre-filter's behavioral signals could disproportionately flag certain demographics. The dwell-time requirement is demographic-neutral (it measures time, not appearance), but the hand-trajectory and object-proximity signals depend on pose estimation accuracy, which may vary across body types and clothing styles. We mitigate this through several design choices:

\begin{itemize}[leftmargin=*]
  \item The pre-filter is intentionally permissive---it flags candidates for VLM review, not for accusation. The VLM provides the actual concealment verdict with natural language justification.
  \item The UNCERTAIN verdict category exists specifically to route ambiguous cases to human review rather than automated accusation.
  \item All alerts include the VLM's natural language description, enabling loss prevention staff to evaluate the reasoning before taking action.
  \item The system is designed as a \textit{decision support tool}, not an autonomous enforcement system. No alert should result in confrontation without human review.
\end{itemize}

We strongly recommend that deployers conduct bias audits on their specific camera setups and customer demographics before operational use.

\section{Discussion}

\subsection{Advantages of Model-Agnostic Orchestration}

The orchestration approach offers several structural advantages beyond cost:

\textbf{Automatic improvement via model swapping.} When a better VLM is released, the system improves by changing two environment variables. For example, an operator currently using Qwen3.5-Omni could switch to Gemma~4 or a future model without any code changes. Trained single-model systems require expensive retraining to incorporate improvements.

\textbf{Explainability.} The VLM provides natural language descriptions of its observations (e.g., ``Person appears to be placing a small object into their right jacket pocket''), which trained classifiers cannot. This is valuable for loss prevention teams who need to understand \textit{why} an alert was generated.

\textbf{Generalization.} VLMs are trained on diverse internet-scale data, making them robust to different store layouts, lighting conditions, and demographics without domain-specific fine-tuning.

\textbf{Transparency.} The source code is available at \url{https://github.com/xHaileab/Paza-AI}, enabling auditing, customization, and community improvement. The pre-filter logic is fully interpretable---operators can inspect exactly why a person was flagged.

\subsection{Limitations}

\textbf{Controlled dataset evaluation.} The DCSASS evaluation uses a synthesized dataset recorded in a controlled laboratory environment. Real retail environments introduce additional challenges including varied lighting, camera angles, occlusion, and more subtle concealment behaviors. The system's built-in logging of every VLM verdict, confidence score, and snapshot provides the infrastructure needed for real-world evaluation during pilot deployments.

\textbf{Latency.} The VLM analysis adds latency after the pre-filter triggers. While alerts still arrive within seconds of the suspicious action, the exact latency is highly dependent on the choice of vision model and hosting infrastructure---self-hosted models on dedicated GPUs achieve 0.5--3\,s, while shared cloud APIs may be slower.

\textbf{VLM dependency.} The system requires a remote GPU or API for VLM inference. Network outages or API failures degrade detection capability, though the retry queue (Section~\ref{sec:retry}) mitigates transient failures.

\textbf{Pre-filter false negatives.} The heuristic pre-filter may miss concealment events that do not trigger any behavioral signal. The pre-filter is intentionally permissive, but some edge cases will be missed.

\textbf{VLM hallucination risk.} VLMs can hallucinate---reporting concealment that did not occur. We employ three mitigations: (1)~\textit{multi-frame consensus}---the VLM must identify consistent evidence across 5 temporally-spaced frames, making it significantly harder to hallucinate a coherent concealment narrative than from a single ambiguous frame; (2)~\textit{structured prompt constraints}---the prompt requires the model to cite specific observable actions rather than vague suspicion, grounding verdicts in evidence; (3)~\textit{confidence thresholds and human review}---the UNCERTAIN category routes ambiguous cases to human review rather than automated alerts. Despite these mitigations, empirical quantification of the false positive rate remains critical future work.

\textbf{Pre-filter parameter sensitivity.} The parameter choices in Table~\ref{tab:params} are based on geometric reasoning and development testing, not formal ablation. Different retail environments (store size, camera angle, product density) may require parameter tuning.

\subsection{Future Work}
\label{sec:future}

\textbf{Large-scale empirical evaluation.} The highest-priority future work is evaluating the system on real retail deployments and larger subsets of the DCSASS dataset. The system already logs every VLM verdict with confidence scores, timestamps, and snapshot images. From pilot deployments, we plan to compute precision, recall, and false positive rates per hour of operation.

\textbf{Pre-filter parameter ablation.} Systematic evaluation of pre-filter parameters ($\tau_d$, $\rho$, $\theta_h$, $\tau_c$) on labeled video to identify optimal values and sensitivity ranges.

\textbf{Self-improving pipeline.} Every VLM verdict can be logged as weakly-labeled training data. After sufficient deployment, a lightweight fine-tuned action classifier could replace the VLM for common cases, further reducing cost while maintaining VLM as a fallback.

\textbf{Edge VLM deployment.} Smaller VLMs (e.g., quantized Gemma~4) may enable on-device inference, eliminating the need for remote GPU infrastructure entirely.

\textbf{Mobile-first SaaS deployment.} We envision a subscription service where store operators need only a camera---all computation runs on centralized GPU infrastructure, with alerts delivered via mobile app.

\textbf{Benchmark creation.} We advocate for the creation of a public retail theft detection benchmark (with appropriate privacy protections) to enable rigorous comparison of detection systems.

\section{Conclusion}

We presented Paza, a zero-shot, model-agnostic retail theft detection framework that orchestrates existing vision models to detect item concealment during shopping sessions without any task-specific model training. Our layered architecture---combining cheap continuous detection (YOLO + ByteTrack + Pose), a multi-signal suspicion pre-filter that reduces VLM calls by 240$\times$, per-person frame buffering for temporal context, and selective multi-frame VLM analysis with structured verdict prompts---demonstrates that model orchestration can serve as a practical and cost-effective alternative to the dominant single-model training approach used by commercial systems.

The system's model-agnostic design ensures it automatically benefits from the rapid pace of VLM improvement: operators can swap between Gemma~4, Qwen3.5-Omni, GPT-4o, or future models without code changes. A capable VLM running on a single RTX 4090 can serve 10--20 stores simultaneously at \$50--100/month per store---a 3--10$\times$ reduction compared to commercial alternatives that require years of proprietary training data.

Preliminary evaluation on the DCSASS shoplifting dataset demonstrates that modern VLMs can perform zero-shot concealment detection when provided with multi-frame temporal context and structured prompts. The system incorporates privacy-preserving design through face obfuscation and minimal data retention, addressing a critical concern in retail surveillance.

Our work suggests a broader principle: for many applied computer vision tasks, the combination of cheap specialized models with selective expensive reasoning models may be more practical than training end-to-end systems---particularly when training data is scarce, expensive, or proprietary. The prompt engineering that transforms a general-purpose VLM into a domain-specific detector is itself a form of ``zero-shot transfer'' that becomes more powerful as VLMs improve. The Paza source code is available at \url{https://github.com/xHaileab/Paza-AI} for reproduction and extension.

\bibliographystyle{plain}

\end{document}